\documentclass[universe,article,accept,pdftex,moreauthors]{Definitions/mdpi}
\firstpage{1} 
\pubvolume{1}
\issuenum{1}
\articlenumber{0}
\pubyear{2026}
\copyrightyear{2026}
\externaleditor{Firstname Lastname} 
\datereceived{3 May 2026} 
\daterevised{1 June 2026} 
\dateaccepted{7 June 2026} 
\datepublished{ }

\usepackage{xcolor}
\Title{A Multi-Level Validation and Traceability Framework for AI-Generated Telescope Scheduling Decisions}

\Author{Hengchu 
 Xiao $^{1,2}$\orcidA{} and Chuanjun Wang $^{1,2,3,4,5,}$*\orcidB{} } 

\AuthorNames{Hengchu Xiao, Chuanjun Wang }

\address{%
$^{1}$ \quad Yunnan Observatories, Chinese Academy of Sciences, Kunming 650216, China; xiaohengchu@ynao.ac.cn
\\
$^{2}$ \quad University of Chinese Academy of Sciences, Beijing 100049, China\\
$^{3}$ \quad Key Laboratory of the Structure and Evolution of Celestial Objects, Chinese Academy of Sciences, \linebreak  {Kunming} 
 650216, China\\
$^{4}$ \quad Yunnan Key Laboratory of Solar Physics and Space Science, Kunming 650216, China\\
$^{5}$ \quad Center for Astronomical Mega-Science, Chinese Academy of Sciences, 20A Datun Road, Chaoyang District, Beijing 100012, China}

\corres{{Correspondence: } 
wcj@ynao.ac.cn}

\abstract{With the gradual introduction of AI into telescope scheduling, AI-based decision-making has shown advantages in handling complex multi-constraint problems. However, its outputs often suffer from inconsistent data references, reasoning errors, and non-executable 
 decisions, limiting applicability in high-reliability observational tasks. In this work, we propose a multi-level validation and traceable reasoning framework that performs systematic reliability verification of AI-generated decisions prior to execution, and enables explicit representation of the reasoning process to support traceable decision-making. The framework integrates data reference validation, logical consistency checks, and observational and instrumental constraint verification to filter and correct invalid decisions. It also introduces atomic reasoning units and their dependency relationships, representing scheduling decisions as a sequence of interconnected reasoning steps that support error localization and post hoc analysis. Experiments show that the framework improves executability and reliability of AI scheduling and reduces loss of transient opportunities. In particular, feedback correction and  {{structured} validation of reasoning steps} enhance the ability to repair and block erroneous decisions, especially in complex scenarios. Compared with pure AI methods, the framework-enhanced approach maintains flexibility while {{substantially} improving reliability and executability}. These results demonstrate a feasible and verifiable pathway for applying AI to high-reliability astronomical observation scheduling.}


\begin{document}


\section{{Introduction} 
}

With the rapid development of time-domain astronomy, observational demands such as large-scale sky surveys and rapid follow-up of transient sources have placed increasingly stringent requirements on telescope scheduling systems. These systems are expected to maximize scientific output under multi-target, multi-constraint, and~dynamically changing conditions, while minimizing the loss of observation opportunities for high-value targets.
The Legacy Survey of Space and Time (LSST) \citep{2019ApJ...873..111I} is expected to generate millions of alerts per night, making target selection from massive alert streams highly complex. Astronomers must manage heterogeneous alert streams across multiple platforms and perform automated scheduling and follow-up confirmation within very short time scales. At~present, the~growth rate of spectroscopic observation resources is significantly lower than that of transient detections. Efficiently allocating limited observational resources to high-value targets, so as to maximize scientific return, has therefore become a key task in telescope and telescope array scheduling in the context of time-domain astronomy. {For example, when a transient alert is issued by a survey such as ZTF, the~scheduler must decide which telescope (and which instrument suite) in a heterogeneous follow-up network such as GROWTH or LCOGT is best suited to perform the observation. In~the case of single-telescope scheduling, the~decision involves evaluating the target's visibility window, observing conditions, scientific priority, instrument availability, and~exposure time to determine whether to interrupt the current task and which instrument suite to use.}
Large-scale survey systems such as Pan-STARRS \citep{2020ApJS..251....3M} and the Mini-SiTian project \citep{2025RAA....25d4002H} demonstrate that wide-field surveys can be efficiently carried out using multiple small telescopes. Telescopes distributed worldwide differ in aperture, detector sensitivity, environmental conditions, target brightness, and~scientific objectives. Deriving a globally optimal and efficient scheduling strategy under such heterogeneous conditions remains the ultimate goal of existing methods.{In this work, “telescope array” refers to a distributed collection of potentially heterogeneous telescopes (different apertures, instruments, and~sites), not to a radio‑interferometric array of identical elements.}

In recent years, significant progress has been made in telescope array scheduling algorithms. For~example, \citet{2019AJ....157..151N} proposed a feature-based automated scheduling method, demonstrating the feasibility of automated scheduling in telescope observations. \citet{2023AJ....165...77Z} developed a distributed multi-layer scheduling framework for wide-field time-domain surveys, which balances global coordination and local flexibility at different levels of granularity. \citet{2023A&C....4400732J} introduced reinforcement learning with reward and penalty mechanisms to improve the self-learning capability of scheduling systems. \citet{2024RAA....24a5003C} applied greedy algorithms to large-scale survey scheduling problems. Building on hierarchical scheduling approaches, \citet{2024AJ....168..214Z} further proposed graph-based reinforcement learning methods, using graph neural networks to enhance coordination among telescopes and improve scheduling performance. \citet{2025RAA....25a5008J} introduced an integer programming-based scheduling framework to reduce time delays between relay observation nodes, thereby improving overall observational~efficiency.

However, time-domain survey scheduling remains a highly complex and dynamic problem that requires real-time decision-making for rapid resource allocation to maximize scientific output. To~address this challenge, previous studies have continuously introduced more sophisticated network structures and adaptive reinforcement learning methods to enhance the global optimization capability of scheduling decisions.
Large-scale time-domain surveys are influenced by numerous factors, and~the scheduling process must account for multiple dynamic constraints, including target priority, visibility windows, instrument status, and~environmental conditions. Therefore, integrating complex model structures with adaptive learning strategies has become an important direction for improving global decision-making and achieving potentially optimal scheduling~performance.

Large language models (LLMs), with~their strong semantic understanding capabilities, provide new solutions for multi-objective trade-offs and decision-making under multiple potential constraints in complex, multi-source alert scenarios. \citet{2025AJ....170...88C} proposed combining deep neural networks with the REINFORCE algorithm to improve the global decision-making capability of models. In~the transient detection and classification pipeline of the Vera C. Rubin Observatory, a~wide range of artificial intelligence tools have been extensively employed \citep{2025NatAs...9.1764R}. \citet{2024Univ...10..210H} presented a comprehensive review of AI applications in optical telescopes, highlighting its advantages in telescope status monitoring and observation decision optimization, and~suggesting that these areas may become important directions for future research.
However, the~misuse of AI technologies has also raised concerns within the astronomical community. \citet{2025NatAs...9.1748T} warned that indiscriminate use of AI may threaten the foundations of academic research. Compared with traditional algorithms, AI reasoning without explicit formal constraints makes it difficult to eliminate concerns about the reliability of its outputs (see Table~\ref{tab:reliability_issues}). As~a result, AI-based approaches have not yet been directly and effectively adopted in high-reliability observational~systems.

\begin{table}[H]
	\caption{{\textls[-25]{{Taxonomy} 
 {of } 
AI failure modes in telescope scheduling and corresponding validation~mechanisms.}}}
	\label{tab:reliability_issues}
	\begin{tabularx}{\textwidth}{CCCC}
		\toprule
		\textbf{Failure Mode} & \textbf{Manifestation} & \textbf{Potential Consequence} & \textbf{Validation Mechanism} \\
		\midrule
		Reference errors 
		& Incorrect or fabricated references to targets, priorities, or~time windows 
		& Wrong targets selected for observation 
		& Raw data consistency~validation \\
		\midrule
		Constraint inconsistency 
		& Omission or contradiction of global constraints (instrument, weather, time windows) 
		& Non-executable schedules 
		& Multi-level constraint~validation \\
		\midrule
		Invalid coordinates/\linebreak  specifications 
		& Physically impossible target positions or instrument configurations 
		& Waste of observational resources 
		& Astronomical and instrumental constraint~validation \\
		\midrule
		Reasoning leaps 
		& Unsupported causal claims or missing evidence links 
		& Untrustworthy decisions 
		& ARU/DAG dependency~checking \\
		\midrule
		Fabricated interference sources 
		& Model-generated disturbances with no basis in input data 
		& Spurious scheduling constraints 
		& Raw data authenticity~validation \\
		\bottomrule
	\end{tabularx}
\end{table}

At present, research on LLMs in astronomy shows a certain imbalance. Most efforts focus on improving the intrinsic capabilities of the models, while insufficient attention is paid to the reliability of model outputs before and after generation. The~inherent ``hallucination'' phenomenon of LLMs prevents their reliability from being adequately guaranteed in critical tasks, and~has gradually become a major bottleneck in the intelligentization of observational scheduling. This significantly limits the large-scale application of such methods in rigorous astronomical scenarios.
Current effective applications are mainly concentrated in educational and research contexts, such as astronomy-oriented LLMs trained on large collections of scientific literature \citep{2025NatSR..1513751D}. \citet{SJPD28F14078CD7811FD7A5346B20E61307A} explored the use of LLMs in wide-field time-domain survey scheduling. However, existing integrations of LLMs in astronomy have not explicitly addressed the reliability issues of AI outputs. As~a result, {LLM-based approaches have not yet been widely adopted in high-reliability operational settings due to reliability concerns, which motivates our validation framework.}

Based on the considerations above, to~address the reliability limitations of AI-driven telescope scheduling, we propose a framework that combines multi-level validation with a traceable reasoning representation. 
On the one hand, the~framework performs systematic verification of executability, consistency, and~astronomical and instrumental constraints for scheduling decisions prior to observation, thereby preventing non-executable scheduling plans from being carried out. On~the other hand, the~basis of scheduling decisions is organized into a traceable and structured reasoning representation, which makes the decision process auditable and supports error localization and result verification.
With these components, the~proposed framework improves the reliability of AI-generated scheduling results in real observational environments while preserving the flexibility of AI-based~scheduling.

\section{Reliability Validation~Framework}
\subsection{Overall Design and~Workflow}

\begin{figure}[H]
    \includegraphics[width=0.85\columnwidth]{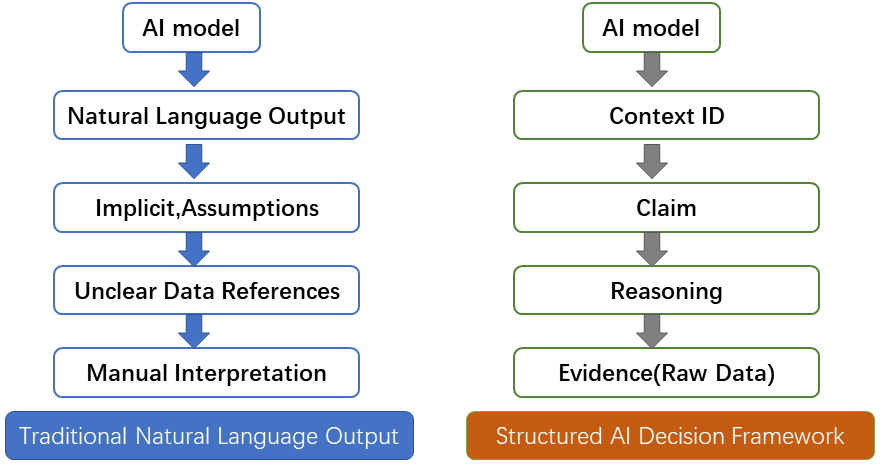}
    \caption{{Comparison} 
 {between} 
 traditional and structured AI scheduling decision~outputs.}
    \label{fig:framework}
\end{figure}

\begin{figure}[H]
    \includegraphics[width= 0.99\columnwidth]{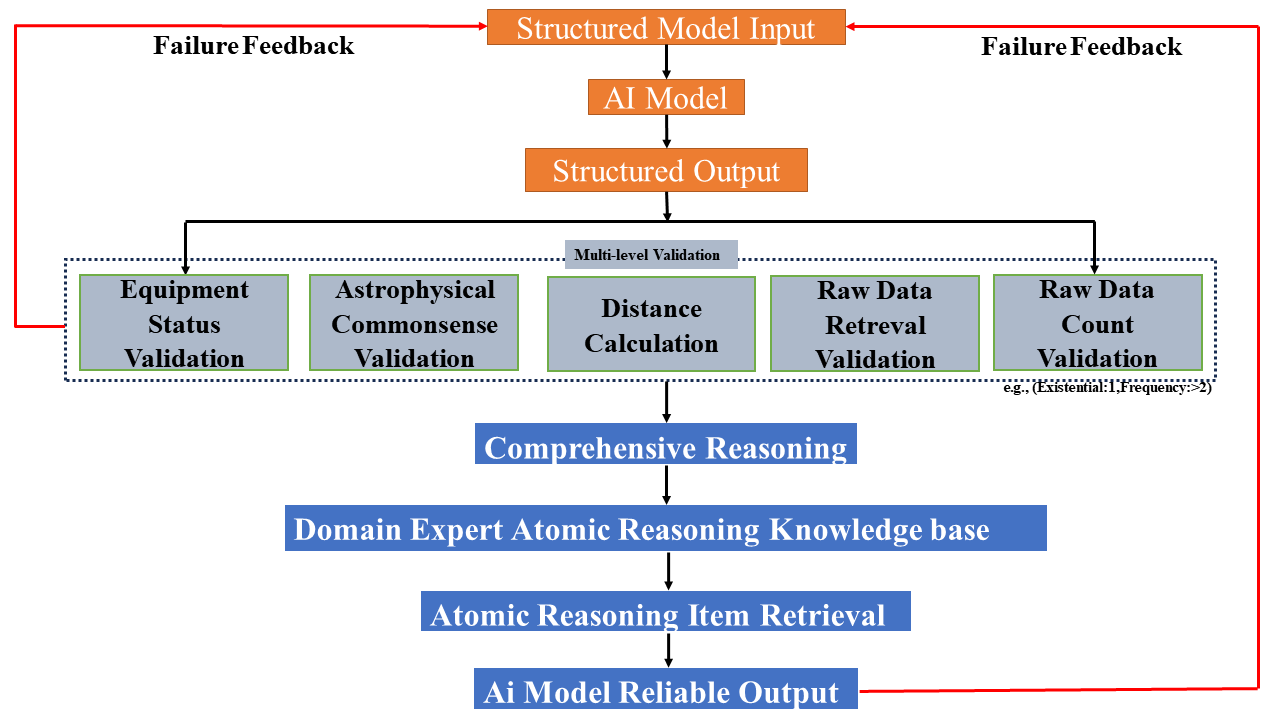}
    \caption{{Flowchart} of the hierarchical verification and visualization decision~framework.}
    \label{fig:workflow}
\end{figure}

The proposed multi-level validation and traceable reasoning framework is positioned between AI-generated scheduling proposals and the execution of observational commands. Without~intervening in the AI decision generation process, the~framework redesigns the model input and output representations based on the original reasoning structure of claims, arguments, and~supporting evidence (see Figure~\ref{fig:framework}). It then performs pre-execution evaluation of the executability and data consistency of the generated scheduling proposals, and~provides executable validation with respect to independence, traceability, reusability, and~rapid correctability of the AI-generated decisions. Figure~\ref{fig:workflow} presents the overall workflow of the proposed framework. In~the overall process, the~LLM first generates scheduling proposals based on structured inputs, including current alert information, instrument status, scientific objectives, and~internal data representations. The~generated proposals are then fed into the validation framework, where they are subjected to multi-level consistency checks and atomic reasoning verification. Only proposals that pass all validation stages are forwarded to the subsequent execution process, while those that fail are blocked. When user feedback is enabled, the~validation results of failed proposals are returned to the AI input stage to support subsequent~refinement.

The framework focuses on determining whether the generated scheduling proposals are logically consistent and physically executable under current observational, instrumental, and~environmental conditions. In~other words, it evaluates the correctness of the output rather than its optimality. The~optimization of scheduling strategies remains the responsibility of the model itself, while the validation framework serves as an independent verification layer.
Furthermore, historical validation results are used to construct high-quality training data, forming a closed-loop interaction between the AI model and the validation process. Each model output is mapped into a structured reasoning representation, where the reasoning steps are decomposed into atomic causal chains and stored for subsequent reuse and arbitration. In~this way, the~system maintains the flexibility of AI-based decision-making while enabling executable validation, feedback-driven correction, and~systematic auditing of the reasoning~process.

\subsection{Structured Representation of Scheduling~Decisions}

AI-generated scheduling decisions are typically expressed in natural language, which is not suitable for automated validation. Therefore, they are transformed into a structured representation to support consistency and constraint verification. This representation retains only the information necessary for~validation.

A complete scheduling proposal is structured into claims, arguments, and~supporting evidence, following the conventional logical process used in human reasoning and problem diagnosis. By~performing automated validation and database-based verification across these levels, the~framework improves the ability to detect issues such as missing global constraints, inconsistent data references, and~conflicts between scheduling decisions and observational or instrumental~conditions.

The structured representation of a scheduling proposal in this framework consists of the following elements: (1) a unique identifier representing the scheduling task; (2) the scheduling decision or operational instruction generated by the AI; (3) the reasoning or justification that links data to the decision; and (4) the data sources or observational conditions supporting the decision. By~explicitly representing these elements, key dependencies in the scheduling proposal are transformed from implicit information into accessible and analyzable structures. As~a result, the~validation process no longer relies on subjective interpretation of natural language outputs, but~instead operates on explicit data references and causal~relationships.

Another objective of this representation is to reduce the complexity of validation, rather than to constrain the model's generation process or expressive flexibility. The~specific organization of fields may vary depending on system implementation, but~this does not affect the validation methodology proposed in this~work.

\subsection{Reasoning Consistency and Constraint~Validation}

This section describes the core mechanisms of the multi-level validation framework, including hierarchical checks on data reference consistency, logical consistency of reasoning chains, and~compliance with astronomical and instrumental constraints in AI-generated outputs. The~validation process determines whether a scheduling decision is consistent and physically executable under current observational conditions and system states, thereby preventing non-executable decisions from affecting the final scheduling process.
As illustrated in Figure~\ref{fig:workflow}, the~layered framework divides the validation procedure into several components, including instrument status validation, astronomical knowledge validation, celestial position and angular distance verification, raw data consistency checks, and~raw data completeness verification. These components incorporate domain knowledge from traditional astronomical scheduling to cover the key aspects of~validation.

\subsubsection{Generation Constraints and Formal Definition of Atomic Reasoning~Units}

The natural language outputs of LLMs typically mix conclusions, supporting evidence, and~implicit assumptions, which can introduce ambiguity and increase the risk of missed detections when directly parsed, thereby affecting the reliability of validation. To~enable verifiable and traceable outputs for AI-based telescope scheduling, a~structured constraint is imposed on the model's reasoning output. This encourages convergence in natural language expressions for similar reasoning processes. Specifically, when generating scheduling decisions, the~model is required to produce a corresponding set of Atomic Reasoning Units (ARUs), which serve as structured inputs for the reasoning consistency validation layer.
An ARU is defined as the smallest indivisible causal reasoning unit to be verified. The~validation framework does not assume its correctness a priori. Logical consistency of the reasoning process is established only when all associated ARUs are individually validated. The~validation process does not depend on matching a complete reasoning chain from historical records. Instead, it operates at the level of individual ARUs, thereby avoiding inconsistencies caused by variations in reasoning order or expression, which would otherwise hinder the recognition of logically equivalent reasoning~paths.

Each ARU is represented as a {tuple:} 
\begin{equation}
a_i = \langle e_i, c_i \rangle
\end{equation}
where $e_i$ denotes the set of data fields supporting the conclusion, such as target priority, visibility window, and~instrument status. The~minimal information unit in $e_i$ must include both the associated table structure and field identifiers. The~component $c_i$ represents the atomic conclusion, for~example: executing observation action $A$ on target $T$ within the time window $[t_1, t_2]$ at node $k$.
The elements in $c_i$ that correspond to telescope state variables or predefined table fields are expressed using a predefined telescope scheduling state dictionary. This constraint promotes convergence in the representation of semantically equivalent ARUs. Typical expressions of telescope state constraints include: \texttt{action\_wait}, \texttt{action\_adjust\_exptime}, \texttt{action\_change\_target}, \texttt{action\_keep\_plan}, \texttt{target\_match}, and~\texttt{interference\_pollution}. By~constraining both representation and data references, the~variability of model outputs for similar actions is~reduced.

At the application level, scheduling states, available resources, and~target attributes are provided to the model in a JSON format aligned with database field names. In~addition, a~telescope scheduling state dictionary and field-level semantic constraints are included in the system prompt. This design encourages the model to produce outputs closer to interface-level descriptions rather than open-ended natural language, thereby reducing variability in expressing equivalent reasoning.
Experimental results show that even for models with tens of billions of parameters, the~number of reusable ARUs for identical reasoning remains on the order of hundreds. This significantly reduces the complexity of reasoning reuse to a manageable level. Under~structured input constraints, most key reasoning components can be consistently mapped to predefined state dictionaries and database fields. For~a small portion of reasoning that cannot be fully structured, the~validation framework adopts a conservative strategy, allowing limited free-text expressions. Experiments indicate that this has only a minor impact on the reuse rate of reasoning~units.

\subsubsection{Construction of Reasoning Dependency DAG and Traceable~Representation}

To represent the dependencies among atomic reasoning units (e.g., a~conclusion depending on visibility calculations or instrument status checks), a~complete scheduling proposal is modeled as a reasoning dependency graph:
\begin{equation}
	G = (V, E)
\end{equation}
where $V = \{a_i\}$ denotes the set of atomic reasoning unit nodes, and~$E \subseteq V \times V$ represents the set of directed dependency edges. If~an edge $a_j \rightarrow a_i$ exists, it indicates that the validity of $a_i$ depends on the conclusion or evidence provided by $a_j$. The~graph is constrained to be acyclic, which allows hierarchical validation along topological paths and enables clear identification of erroneous atomic reasoning units when validation~fails.

\begin{figure}[H]
    \includegraphics[width=\columnwidth]{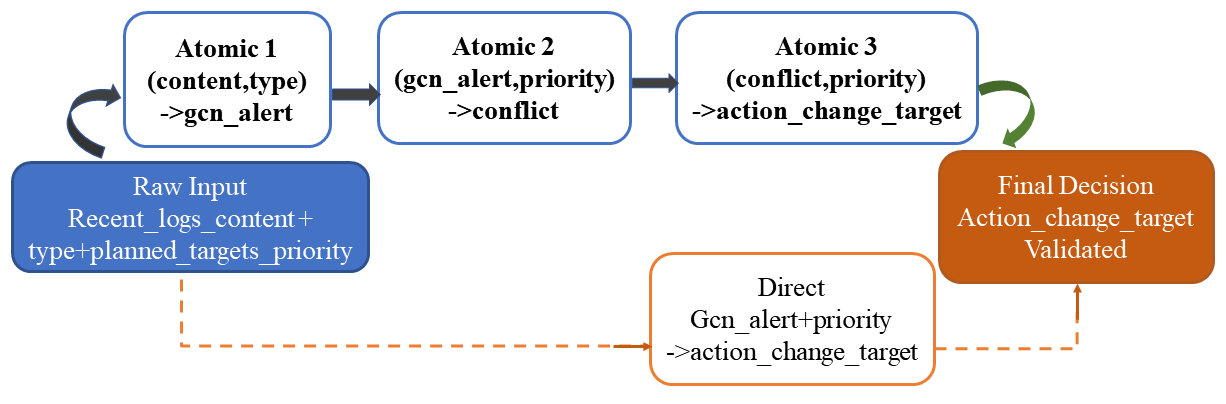}
    \caption{{Final} decision reasoning via atomic reasoning chain~construction.}
    \label{fig:fig3}
\end{figure}

The introduction of the DAG serves three purposes. First, the~validation process can perform local blocking based on dependency relationships, preventing high-cost constraint evaluations when prerequisite conditions are not satisfied. Second, validated subsets of atomic assertions can be stored as reusable reasoning units in a knowledge base for auditing and reuse. Since atomic reasoning units are inherently interpretable, manual inspection, modification, and~extension of reasoning components become feasible, achieving a traceable and structured representation of the decision process (see Figure~\ref{fig:fig3}). Third, the~framework allows the prediction of final outputs by exploring possible topological combinations of ARUs under given~inputs.

Experimental results further illustrate how complete reasoning paths are constructed through chained DAG structures of atomic reasoning units. As~shown in Figure~\ref{fig:fig3}, one example of a frequently occurring reasoning chain is visualized through DAG-based path tracing. The~reasoning paths from input to output can vary significantly; even when two reasoning processes are semantically equivalent, the~model may traverse different paths to produce the same final decision.
By combining ARUs through DAG-based structures within the knowledge base, it becomes possible to anticipate potential reasoning paths for a given input. These paths form a visual representation of the reasoning process. Logical validation of reasoning is performed by evaluating the executability of these paths. A~valid reasoning chain requires that all constituent ARUs can be retrieved from the expert knowledge base, ensuring correctness at the atomic level while preserving diversity in model-generated~expressions.

{To summarize, the~validation mechanisms described in this section constitute a form of ``step-level consistency checking''. It is important to precisely define its scope. This mechanism does not verify the global correctness or semantic validity of the entire reasoning chain. Instead, it operates as a local verification at the level of individual reasoning steps: it checks whether each atomic reasoning unit (ARU) is supported by consistent data references, and~whether the dependencies between ARUs form a valid acyclic graph. In~this sense, step-level consistency checking provides a fine-grained, local validation of the reasoning structure—it ensures that the building blocks and their connections are sound—without making claims about the global optimality or ultimate truth of the final~decision.}

\subsubsection{Failure Blocking and Closed-Loop Feedback~Mechanism}

Scheduling proposals that fail validation are not allowed to enter the downstream observation control pipeline. For~each failed case, the~validation framework outputs structured error information, including the identifier of the failed validation item, the~type of failure (data inconsistency, logical inconsistency, or~violation of hard constraints), the~associated fields and constraint conditions, and~the corresponding traceable dependency paths.
For validation failures such as missing fields, time window conflicts, incorrect target references, physically non-visible targets, or~unavailable instruments, the~structured error information can be fed back to the model input as additional constraints. This feedback guides the model to generate revised candidate proposals that satisfy the required conditions. Historical records generated through this closed-loop feedback process can further serve as high-quality datasets for subsequent model refinement.
In this way, the~validation framework preserves the flexibility of AI-based decision-making while ensuring execution safety within deterministic constraint checks, and~establishes a closed-loop process that supports rapid error~correction.

\section{Experimental~Validation}
\unskip

\subsection{Experimental~Design}

To evaluate the effectiveness of the proposed multi-level validation and traceable reasoning framework in telescope scheduling, a~unified experimental setup is designed from both scheduling performance and reliability perspectives. At~the scheduling level, observation task scenarios triggered by transient alerts are constructed to compare the performance of traditional scheduling methods, pure AI-based scheduling, and~AI-based scheduling enhanced with the proposed validation framework under complex constraints. At~the reliability level, quantitative analysis is conducted on the effectiveness of each validation stage prior to execution, with~a focus on improvements in constraint satisfaction, consistency, and~execution~safety.

To balance feasibility and practical relevance, two complementary types of scenarios are considered: (1) simulated scheduling scenarios, which are used to analyze the behavioral characteristics and failure modes of AI-based scheduling under typical constraints; and \linebreak  (2) scheduling scenarios constructed from real alert data, where candidate targets are derived from ZTF transient alert data to evaluate the robustness of the framework under realistic data distributions and interference~conditions.

In both types of experiments, telescope scheduling states are represented using a predefined state dictionary. Together with alert information, these are provided to the model in a JSON format aligned with database fields. This design reduces the impact of variability in natural language expressions on reasoning consistency and brings model outputs closer to structured scheduling~interfaces.

Two experimental configurations are adopted:  
(1) Direct execution configuration, in~which AI-generated scheduling proposals are directly evaluated by the execution feasibility validation module;  
(2) Validation-feedback configuration, in~which AI outputs must pass the multi-level validation framework before execution. Proposals that fail validation are blocked, and~the corresponding failure information is returned. When feedback is enabled, the~failure information is fed back to the model input to trigger further reasoning and~correction.

In addition, several traditional scheduling methods are introduced as baselines, including deadline-based rule scheduling, priority-based rule scheduling, and~heuristic scheduling methods.
Through this experimental design, we aim to systematically evaluate: (1) the reliability limitations of AI-based scheduling without validation; and (2) the ability of the proposed framework to improve the executability and consistency of scheduling decisions across different~scenarios.

\subsection{Representative Scheduling~Scenarios}\label{sect:3.2}

We construct three categories of representative transient scheduling scenarios to cover common structures such as interrupt-driven scheduling, priority conflicts, and~multi-target competition. These are further divided into six specific test scenarios:

S1: Single high-priority transient scenario. The~system is in routine observation mode when a high-priority alert is triggered. The~task is to complete follow-up observations within a limited visibility~window.

S2: Single low-priority transient scenario. The~system is executing higher-priority routine tasks when a low-priority alert arrives. This scenario is used to evaluate scheduling consistency under priority~conflicts.

S3: Multiple simultaneous transient alerts (multi-priority competition). Multiple alerts are triggered simultaneously. This scenario examines whether AI-based scheduling is more prone to reference inconsistencies, conflicts, or~missing constraints under multi-target and multi-constraint~conditions.

S4: Low-priority ToO scenario with long visibility windows under multiple irrelevant or outdated interference signals. This scenario evaluates the robustness of model outputs in the presence of significant noise and~interference.

S5: High-priority ToO scenario in complex environments. Dozens of irrelevant or outdated environmental inputs, as~well as recent environmental changes, are introduced. Under~multiple high-priority alerts, this scenario examines issues such as incorrect references, missing reasoning evidence, and~violations of physical~constraints.

S6: Abnormal device and out-of-bound physical constraint scenario. This includes cases with physically invalid celestial coordinates (e.g., RA/Dec beyond valid ranges) and abnormal instrument configurations (e.g., invalid filters), simulating model behavior when inputs exceed its capability~limits.

\subsubsection*{{Benchmark} 
 Construction and Statistical~Treatment}

{
Each scenario was implemented as a statistical study comprising multiple independently repeated test instances (i.e., the~same test template was run many times), rather than a single fixed test case. }Table~\ref{tab:validation_phase} hl{reports, for~each model and scenario, the~total number of initial validation tests and the number that passed all validation checks. These numbers accumulated gradually through repeated runs; they were not a pre‑specified target. The~overall total (2433) corresponds to the ``more than 2000'' tests mentioned in the text. The~final ablation study} (Table~\ref{tab:table3}) hl{uses a balanced set of 60 cases per scenario \linebreak  (360 total) and reports 95\% confidence intervals to quantify statistical uncertainty.}

\begin{table}[H]
	\caption{Initial validation tests (without feedback) across the six scenarios. For~each model, the~table reports the total number of tests and the number of tests that passed all validation checks. Each scenario was implemented as a statistical study comprising multiple independently repeated test instances (i.e., the~same test template was run many times). Total tests exceed 2000, as~referenced in the~text.}
	\label{tab:validation_phase}
	\begin{tabularx}\textwidth{cCCCCc}
		\toprule
		\multirow{2}{*}{\vspace{-5pt}\textbf{Scenario}} & \multicolumn{2}{c}{\textbf{Phi-4 ({14B})}}
		 & \multicolumn{2}{c}{\textbf{Qwen3 (30B)}} & \multirow{2}{*}{\vspace{-5pt}\textbf{Total per Scenario}} \\
		\cmidrule(lr){2-5}
		& \textbf{Total} & \textbf{Passed} & \textbf{Total} & \textbf{Passed} & \\
		\midrule
		S1 & 200 & 195 & 200 & 193 & 400 \\
		S2 & 204 & 203 & 200 & 200 & 404 \\
		S3 & 215 & 212 & 200 & 200 & 415 \\
		S4 & 208 & 193 & 200 & 197 & 408 \\
		S5 & 200 & 170 & 200 & 186 & 400 \\
		S6 & 204 & 69  & 202 & 40  & 406 \\
		\midrule
		Total & 1231 & 1042 & 1202 & 1016 & \textbf{{2433} 
} \\
		\bottomrule
	\end{tabularx}
\end{table}
\vspace{-8pt}

\begin{table}[H]
	\caption{{Performance} 
 comparison of representative scheduling methods (Panel A) and ablation study of the Qwen3-based pipeline (Panel B) on the controlled six-scenario benchmark constructed from historical ZTF alert candidates. Values are reported with 95\% confidence intervals for overall metrics where applicable. Scenario-wise columns report RPA~values.}
	\label{tab:table3}
	\footnotesize
\begin{adjustwidth}{-\extralength}{0cm}

	\begin{tabularx}\fulllength{ccccccccccc}
		\toprule
		\multicolumn{11}{c}{\textbf{Panel A. Main Method Comparison}} \\
		\midrule
		\textbf{Method} & \textbf{EER} & \textbf{RPA} & \textbf{RRS} & \textbf{S1} & \textbf{S2} & \textbf{S3} & \textbf{S4} & \textbf{S5} & \textbf{S6} & \textbf{{Runtime} (s)} 
 \\
		\midrule
		Greedy & {$1.000~[0.989,1]$} 
 & \textls[-15]{$0.769~[0.723, 0.810]$} & \textls[-15]{$0.885~[0.858, 0.909]$} & 1.000 & 1.000 & 0.400 & 0.733 & 0.483 & 1.000 & $<$0.001 \\
		Deadline & $1.000~[0.989,1]$ & \textls[-15]{$0.711~[0.662, 0.756]$} & \textls[-15]{$0.711~[0.664, 0.756]$} & 1.000 & 0.267 & 1.000 & 0.000 & 1.000 & 1.000 & $<$0.001 \\
		ILP & $1.000~[0.989,1]$ & $1.000~[0.989,1]$ & $1.000~[1,1]$ & 1.000 & 1.000 & 1.000 & 1.000 & 1.000 & 1.000 & 0.024 \\
		Qwen3 Only & \textls[-25]{$0.692~[0.642, 0.737]$} & \textls[-15]{$0.531~[0.479, 0.582]$} & \textls[-15]{$0.612~[0.565, 0.657]$} & 1.000 & 1.000 & 0.050 & 0.083 & 0.050 & 1.000 & 174.7 \\
		\textls[-25]{{Qwen3 + Full}} & $1.000~[0.989,1]$ & \textls[-15]{$0.944~[0.916, 0.964]$} & \textls[-15]{$0.983~[0.975, 0.990]$} & 1.000 & 1.000 & 0.733 & 1.000 & 0.933 & 1.000 & 184.3 \\
		\midrule
		\multicolumn{11}{c}{\textbf{Panel B. Qwen3 Ablation}} \\
		\midrule
		\textbf{Config.} & \textbf{EER} & \textbf{RPA} & \textbf{RRS} & \textbf{S3} & \textbf{S4} & \textbf{S5} & \textls[-15]{\textbf{Repair}} & \textls[-15]{\textbf{{R.} Succ}} & \textls[-15]{\textbf{Audit}} & \textls[-15]{\textbf{Runtime (s)}} \\
		\midrule
		Qwen3 Only & \textls[-25]{$0.692~[0.642, 0.737]$} & \textls[-15]{$0.531~[0.479, 0.582]$} & \textls[-15]{$0.612~[0.565, 0.657]$} & 0.050 & 0.083 & 0.050 & 0.000 & -- & -- & 174.7 \\
		+Val & $1.000~[0.989,1]$ & \textls[-15]{$0.683~[0.634, 0.729]$} & \textls[-15]{$0.765~[0.724, 0.804]$} & 0.050 & 1.000 & 0.050 & 0.000 & -- & 0.000 & 174.7 \\
		{+Val + FB} & $1.000~[0.989,1]$ & \textls[-15]{$0.928~[0.896, 0.950]$} & \textls[-15]{$0.978~[0.969, 0.987]$} & 0.650 & 1.000 & 0.917 & 0.308 & 0.523 & 0.000 & 214.3 \\
		+Full & $1.000~[0.989,1]$ & \textls[-15]{$0.944~[0.916, 0.964]$} & \textls[-15]{$0.983~[0.975, 0.990]$} & 0.733 & 1.000 & 0.933 & 0.308 & 0.532 & 0.056 & 184.3 \\
		\bottomrule
	\end{tabularx}
\end{adjustwidth}
	
	\parbox{\textwidth}{\footnotesize{Note}: Scenario columns report RPA. Each scenario: 60 cases, total 360/method. {R. }Succ = Repair Success Rate (conditional on repair attempt).}
\end{table}

\subsection{Validation Categories and Error~Types}

To align with practical system testing procedures, the~validation framework is divided into 5 levels:

\textbf{{Instrument state constraints:}
} Any AI-generated decision must first satisfy constraints related to instrument status and observational conditions. For~example, AI-generated decisions should be disabled when cloud coverage exceeds 90\% or when instruments are~unavailable.

\textbf{{Astronomical knowledge constraints:}} This level checks for violations of basic astronomical observational knowledge, such as invalid filter usage, expired time windows, unreasonable parameter ranges, or~operations that conflict with predefined safety~rules.

\textbf{Computational constraints:} Since AI models are primarily based on probabilistic inference, their reliability in numerical computations may be insufficient. Therefore, key astronomical calculations, such as angular distances between targets, are re-evaluated using traditional methods to ensure~correctness.

\textbf{Raw data authenticity and traceability validation:} This level verifies whether the targets, instrument states, and~constraint fields referenced in scheduling proposals can be resolved and validated against database records or state dictionaries. It is particularly effective in detecting non-existent references or inconsistencies between referenced data and actual system~states.

\textbf{Consistency between decision expressions and raw data:} This level examines whether the language description of decision reasoning is consistent with the amount of underlying referenced data. For~example, if~an interference source is described as frequent, the~corresponding referenced raw data should contain multiple entries; if an interference is claimed to exist, at~least one corresponding data record should be~present.

\subsection{Reasoning~Verification}

\subsubsection*{{Reasoning} Validation and~Visualization}

The reasoning validation process is based on atomic reasoning units and their corresponding directed acyclic graph (DAG) representation. The~reasoning output of the model is decomposed into a set of atomic reasoning units. If~any invalid atomic unit is detected within the reasoning chain, or~if unstable reasoning patterns are observed, the~entire scheduling proposal is considered unacceptable and is~blocked.

The visualization of reasoning relies on tracing the causal DAG of atomic reasoning units. For~a given input, the~reasoning process of AI-generated scheduling decisions can be visualized through DAG-based paths. Users can trace, audit, and~analyze the reasoning outputs by following the DAG composed of atomic reasoning units. Furthermore, the~reasoning knowledge base can be verified and refined based on these structures. Figure~\ref{fig:fig3} presents an example of reasoning visualization for a set of frequently reused atomic reasoning units obtained from experimental results. The~DAG structure demonstrates that the proposed approach for constructing a reasoning knowledge base is feasible. At~the same time, the~diversity of natural language expressions is effectively constrained through structured reasoning~representation.

\subsection{Evaluation~Metrics}

To quantitatively assess the improvement in reliability of AI-assisted telescope scheduling provided by the proposed multi-level validation framework, we adopt statistical metrics aligned with the practical validation process of the system. Since this work focuses on the executability and consistency of scheduling decisions rather than their optimality, the~evaluation emphasizes the proportion of non-executable outputs, the~distribution of failure causes, the~effectiveness of validation and feedback mechanisms, and~the reusability of atomic reasoning~units.

\begin{enumerate}[align=right,leftmargin=7.6mm,labelsep=3.2mm]
\item [\textbf{(1)}] \textbf{{Pass Rate (PR)}
}
\end{enumerate}

The pass rate measures the proportion of AI-generated scheduling proposals that successfully pass validation and proceed to execution. Let the total number of test samples be $N$, and~the number of proposals that pass validation be $N_{\mathrm{pass}}$, then:
\begin{equation}
	\mathrm{PR} = \frac{N_{\mathrm{pass}}}{N}
\end{equation}

{This} 
 metric is computed for both baseline (without validation) and verified (with validation framework) configurations across different scheduling scenarios, reflecting the ability of the framework to filter out non-executable~proposals.

\begin{enumerate}[align=right,leftmargin=7.6mm,labelsep=3.2mm]
\item [\textbf{(2)}] \textbf{{Repair Success Rate (RSR)}}
\end{enumerate}

When the feedback loop (failure feedback followed by model regeneration) is enabled, the~effectiveness of correction can be evaluated. Let $N_{\mathrm{repair},1}$ denote the number of proposals that become valid after one feedback iteration, and~$N_{\mathrm{block}}$ denote the total number of blocked proposals. The~single-step repair success rate is defined as:
\begin{equation}
	\mathrm{RSR}_1 = \frac{N_{\mathrm{repair},1}}{N_{\mathrm{block}}}
\end{equation}

\begin{enumerate}[align=right,leftmargin=7.6mm,labelsep=3.2mm]
\item [\textbf{(3)}] \textbf{{Failure Distribution (FD)}}
\end{enumerate}

To analyze the main sources of non-executable outputs, blocked samples (without feedback correction) are categorized according to validation error types. This reveals the distribution of failure modes in AI-generated scheduling~decisions.

\begin{enumerate}[align=right,leftmargin=7.6mm,labelsep=3.2mm]
\item [\textbf{(4)}] \textbf{{Reasoning Reusability}}
\end{enumerate}

The reusability of reasoning is evaluated based on whether atomic reasoning units can be consistently reused and whether the complexity of reusable reasoning units converges. This metric reflects the effectiveness of the proposed framework in enabling traceable reasoning. Experimental results across different scheduling scenarios are used to assess the reuse capability of reasoning~components.

\subsection{Experimental Results and~Analysis}

\subsubsection{Reliability Evaluation Under Representative Scheduling~Scenarios}

\begin{figure}[H]
    \includegraphics[width=\columnwidth]{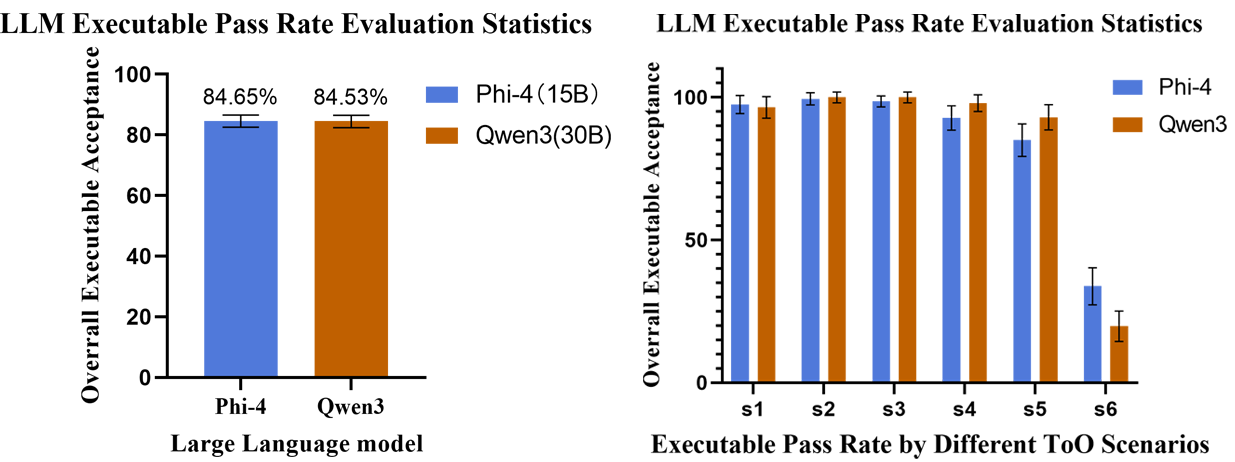}
    \caption{\textls[-15]{Overall {e}
xecutable {p}ass rates of two different models across various ToO scheduling~scenarios.}}
    \label{fig:fig4}
\end{figure}

Across all tested scenarios, the~executability performance of AI-generated scheduling proposals without validation feedback correction is summarized in Figure~\ref{fig:fig4}.

Using the \textit{{microsoft phi-4}
} (14B) model, the~average pass rate across all scenarios is 79.3\% over 1231 test cases. For~the \textit{{Qwen}} (30B) model, the~average pass rate is 84.5\% over 1202 test~cases.

The pass rate analysis across different scenarios shows that models achieve significantly higher executability for simple and short-input scheduling tasks than for complex and long-input multi-ToO follow-up scenarios. 
When telescope operation conditions are abnormal, or~when scheduling targets exceed physically meaningful limits, the~models exhibit limited capability in identifying, correcting, and~filtering erroneous information. In~such cases, the~pass rate reaches its lowest level, indicating the most significant impact on efficient telescope~scheduling.

\begin{figure}[H]
    \includegraphics[width=0.85\columnwidth]{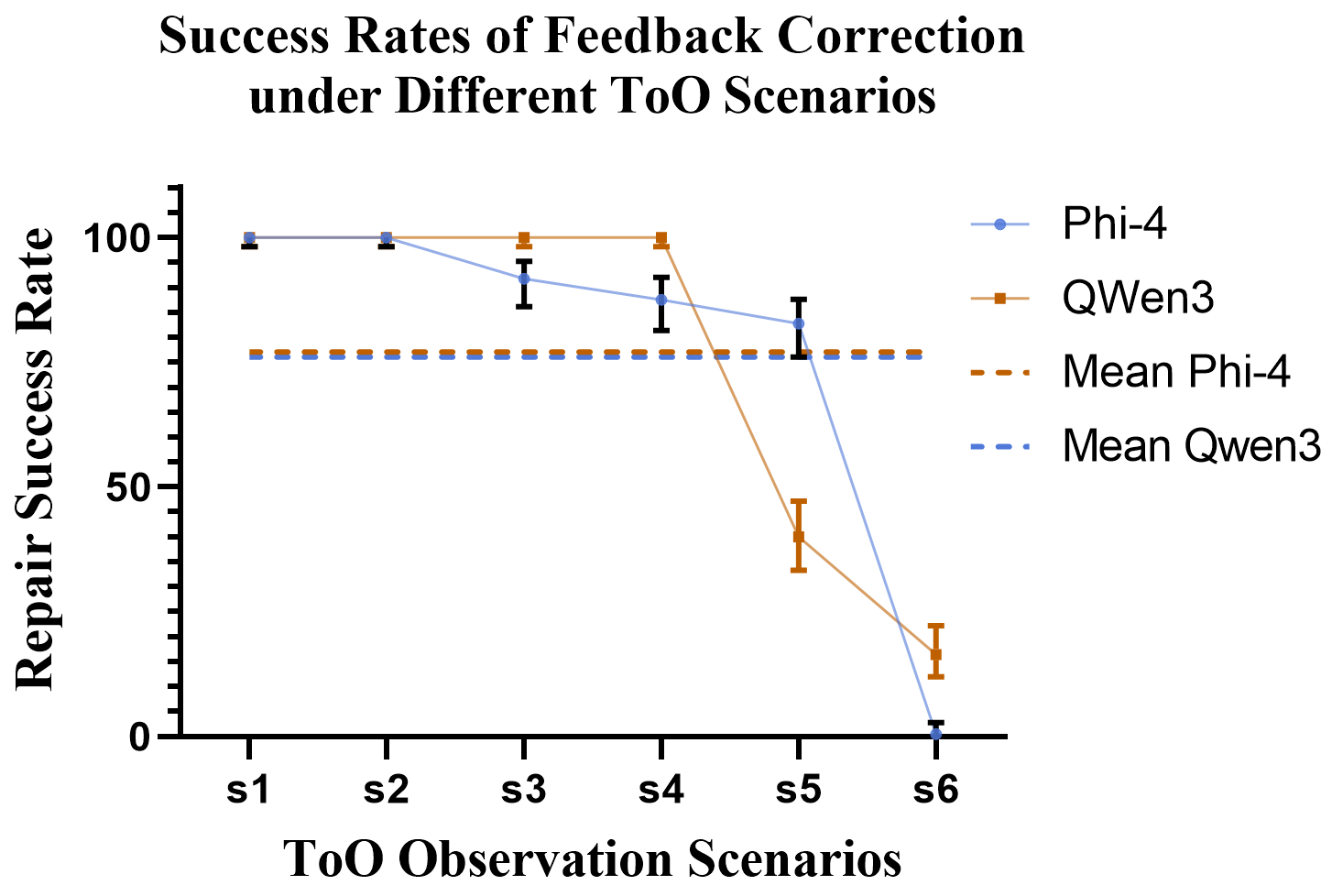}
    \caption{Model {p}erformance {a}fter validation feedback across different ToO observation~scenarios.}
    \label{fig:fig5}
\end{figure}

\textls[-15]{Subsequent experiments focus on automated correction of model outputs through validation feedback. In~different scenarios, validation results are fed back into the model as inputs for iterative correction. The~results show that the validation-feedback mechanism significantly improves the model’s correction capability. Although~the improvement remains effective in complex scenarios, the~degree of correction is reduced (see Figure~\ref{fig:fig5}). At~the same time, the~framework demonstrates strong error detection capability. The~multi-level validation framework substantially improves the reliability of model outputs (see Figure~\ref{fig:fig4}), and~confirms the feasibility of automated validation and feedback-driven correction~mechanisms.}

\begin{figure}[H]
    \includegraphics[width=0.87\columnwidth]{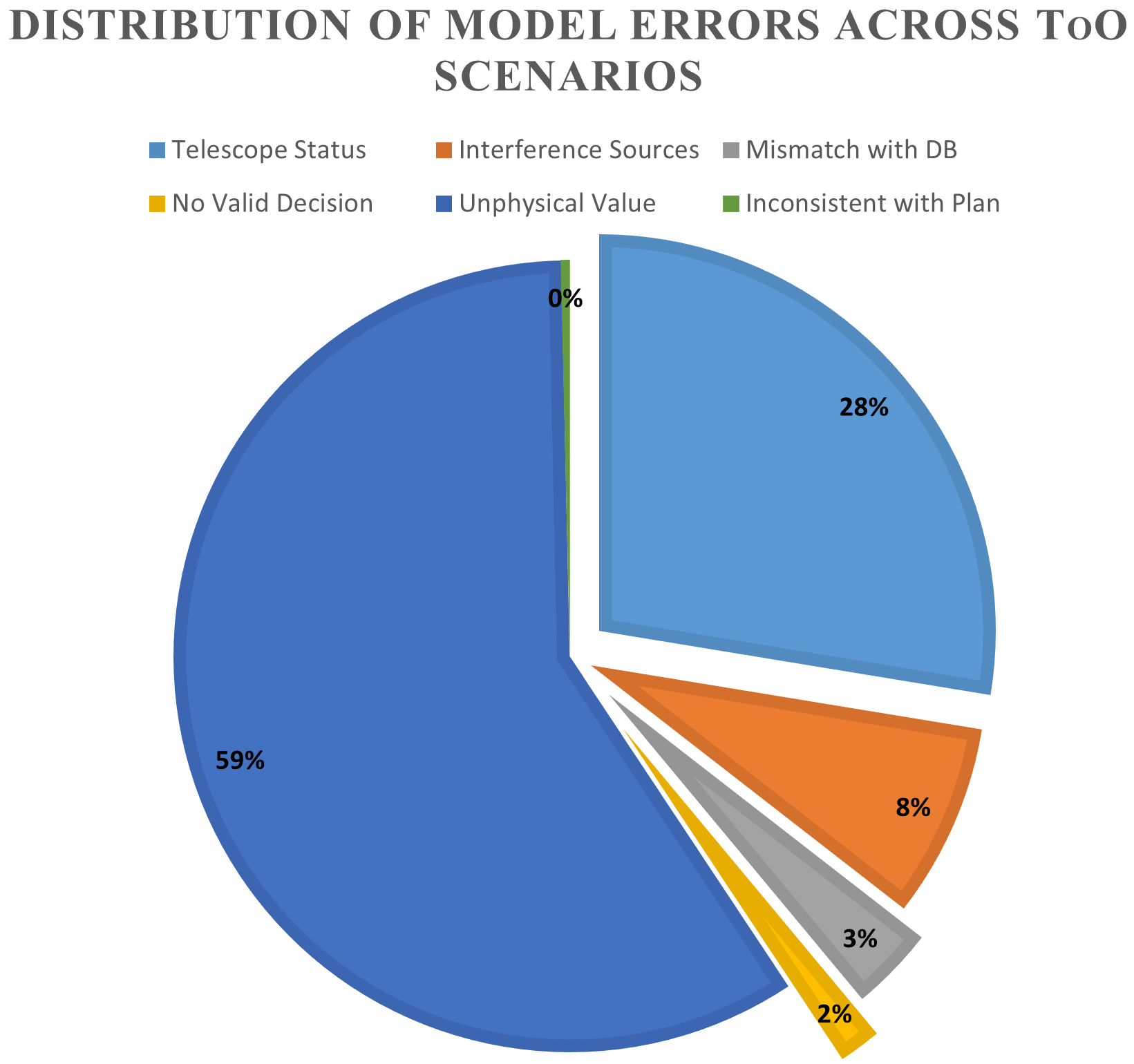}
    \caption{Distribution of {m}odel {o}utput error types across different ToO~scenarios.}
    \label{fig:fig6}
\end{figure}

The statistical analysis of model errors across more than 2000 ToO scheduling tests shows that the most common failure cases can be summarized as follows. First, models exhibit limited capability when handling abnormal physical values or instrument states that exceed reasonable ranges. Second, models tend to omit constraints related to the current telescope state. In~addition, models often struggle to distinguish between the presence of interference and varying levels of interference intensity (see Figure~\ref{fig:fig6}). In~rare cases, models generate spurious interference signals that do not exist in the input data. The~performance of different models varies in this aspect. In~this study, the~\textit{{microsoft phi-4 }} (14B, Microsoft Research, Redmond, WA, USA) model shows fewer such hallucinated cases compared to the \textit{{Qwen 3}} (30B, Qwen Team, Alibaba Group, Hangzhou, China) model. This suggests that hallucination issues can be mitigated through model improvements, but~cannot yet be completely eliminated. This further demonstrates the necessity of the proposed validation framework for multi-level verification and correction of model outputs. In~addition, the~validation results generated by the framework provide high-quality historical data for subsequent model~refinement.

After introducing the hierarchical validation framework, the~executability of AI-generated scheduling proposals shows consistent improvement across all test scenarios. Specifically, compared to the baseline configuration without validation, introducing validation at the output level alone increases the average pass rate by 18.1\%. Building on this, enabling the feedback mechanism—where failed proposals are returned to the model for regeneration—leads to an additional improvement of 12.6\%, indicating that structured feedback effectively guides the model toward generating constraint-compliant decisions.
Furthermore, by~incorporating reasoning traceability and validation based on atomic reasoning units (see Figure~\ref{fig:fig3}), the~average pass rate is further improved by 16.8\%. This improvement is more pronounced in complex scenarios, with~increases of 31.35\% in scenario S4 and 25\% in scenario S5. These results indicate that the proposed mechanism is particularly effective in situations where reasoning errors and constraint conflicts are more likely to occur.
{To clarify the interpretation of these gains, the~staged comparison can be understood as four configurations: (1) AI-only generation; (2) AI generation followed by output-level validation; (3) AI generation with validation and feedback-guided regeneration; and (4) AI generation with validation, feedback, and~ARU/DAG-based reasoning verification. Under~this interpretation, the~reported 18.1\%, 12.6\%, and~16.8\% gains are cumulative and sequential: each percentage refers to the improvement of adding one layer on top of the immediately preceding configuration. Specifically, the~16.8\% gain represents the incremental benefit of adding reasoning traceability and reasoning-level verification (i.e., moving from (3) to (4)). The~improvement is especially visible in complex scenarios, where invalid references and reasoning conflicts are more likely to occur. All four configurations were evaluated uniformly across scenarios S1–S6.}
Overall, the~results demonstrate that the proposed framework improves the executability of AI-assisted scheduling decisions while significantly enhancing the reliability and traceability of the reasoning process. The~framework effectively filters inconsistencies in reasoning and prevents non-executable decisions from entering the execution~stage.

{Scenario S3 involves simultaneous multi-alert competition under coupled constraints, where the model must jointly resolve priority, visibility, and~timing conflicts across multiple targets. In~this scenario, Qwen3 exhibits a notable decline in the quality of its atomic reasoning units: expression normalization weakens, causal dependencies become less rigorous, and~attribution errors increase. These degradations occur specifically under multi-target coupling, suggesting that the model's reasoning capability is strained when constraints must be jointly satisfied across competing alerts. This also illustrates the diagnostic value of step-level consistency checking: by examining individual ARUs rather than only the final decision, one can identify which specific aspects of reasoning deteriorate under complex conditions—information that aggregate metrics alone cannot provide.}

\begin{figure}[H]
    \includegraphics[width=0.9\columnwidth]{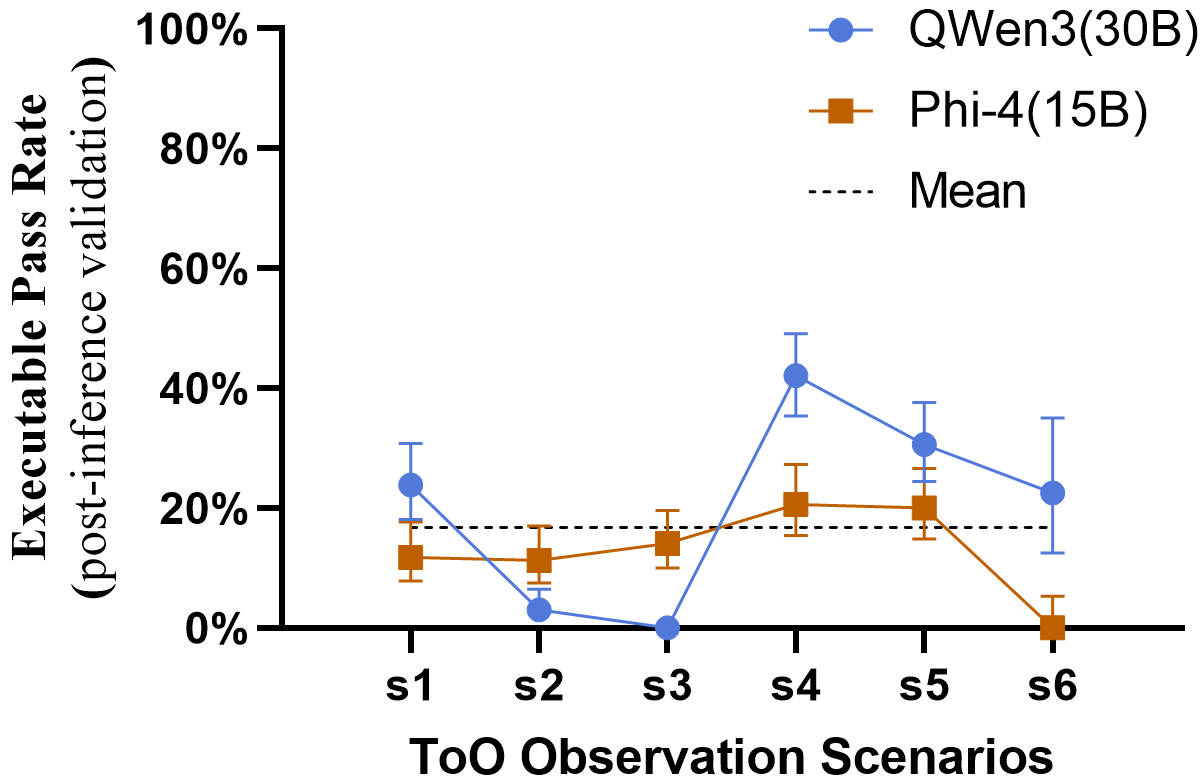}
    \caption{{Post}
-inference-validation executable pass rates for Qwen3 and Phi-4 models under scenarios S1--S6. The~dashed line indicates the~average.}
    \label{fig:fig7}
\end{figure}

\subsubsection{Reasoning Visualization and Interpretability~Analysis}

The previous experiments focus on improving the reliability of model outputs through multi-level validation, primarily by filtering out incorrect references and physically invalid decisions. However, the~auditability of reasoning relies on the subsequent analysis of reasoning visualization.
The effectiveness of AI reasoning visualization depends on two key aspects: (1) whether atomic reasoning units can be reused; and (2) whether the number of distinct expressions for similar reasoning units converges, such that the complexity of reasoning remains within a manageable~range.

\begin{figure}[H]
    \includegraphics[width=0.95\columnwidth]{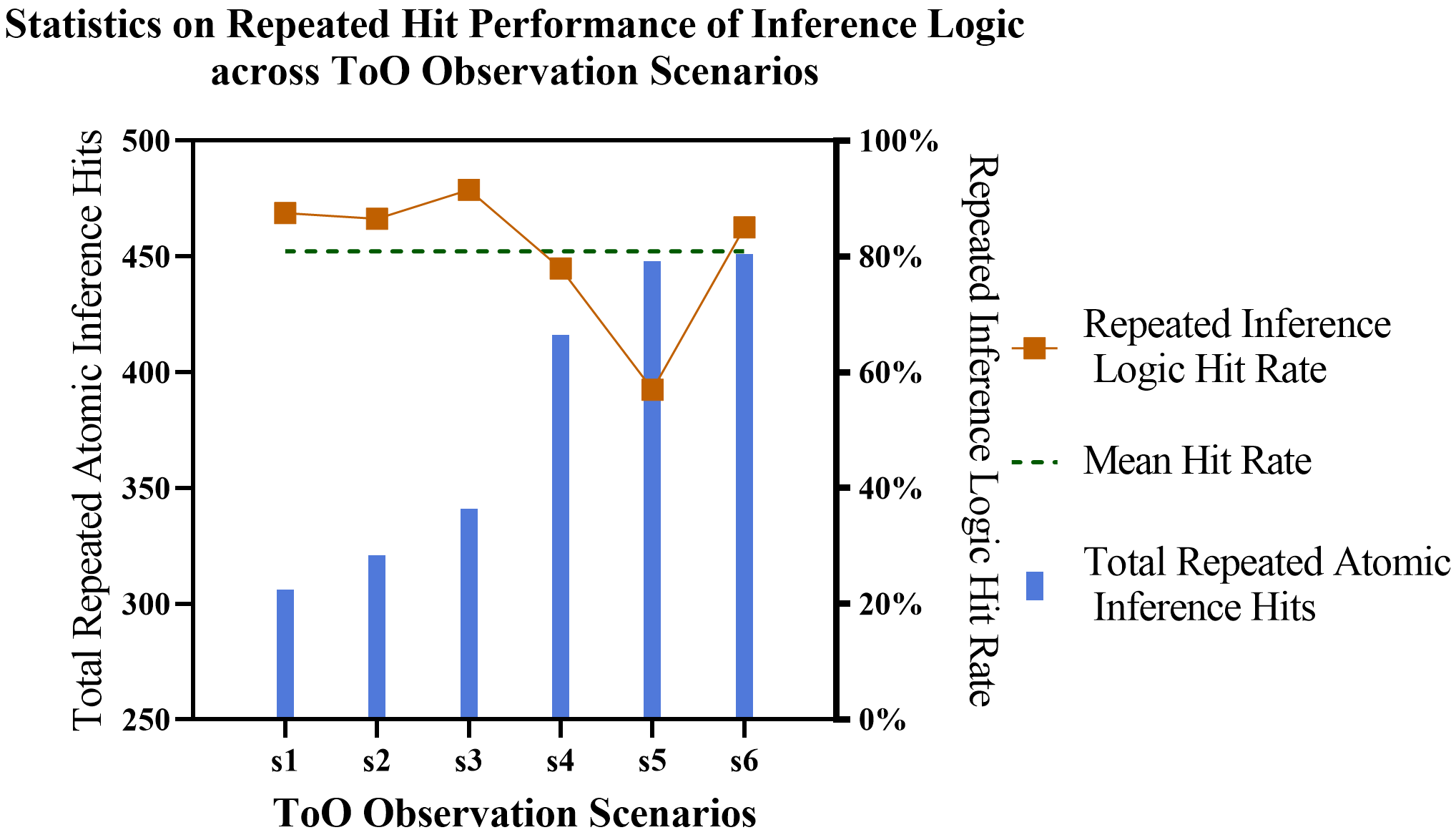}
    \caption{Inference {l}ogic repetition hit rates across different ToO~scenarios.}
    \label{fig:fig8}
\end{figure}

To evaluate the reusability of the proposed explicit reasoning knowledge base and to provide evidence for traceable reasoning in AI-based scheduling, we conduct experiments using the \textit{{Qwen}} (30B) model. Across more than 2000 test cases, when the number of frequently occurring atomic causal reasoning units reaches approximately 300–320, the~average reasoning reuse rate across all scenarios is about 80\% (see Figure~\ref{fig:fig8}).
Among the different scenarios, the~reuse rates reach 87.5\% and 86.5\% in S1 (single high-priority transient scenario) and S2 (single low-priority transient scenario), respectively, while the reuse rate reaches 90\% in S3 (multi-transient scheduling scenario).

\begin{figure}[H]
    \includegraphics[width=\columnwidth]{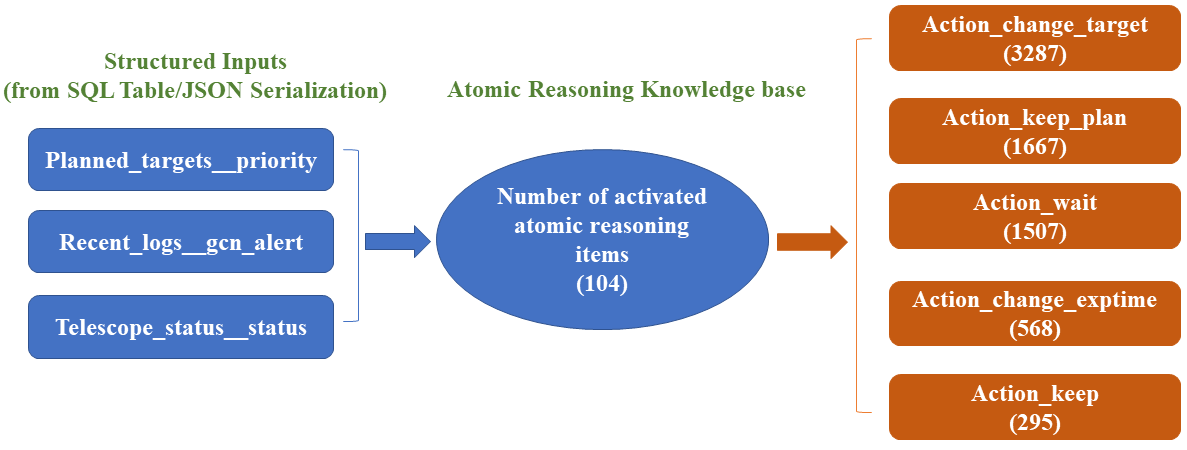}
    \caption{Statistical visualization of reasoning based on model inference results. The~numbers in parentheses indicate the actual counts recorded during the model testing experiments. The~figure illustrates the distribution of reasoning paths and the activation of atomic reasoning units under structured inputs, demonstrating that a large number of reasoning paths can be represented by a limited set of reusable atomic reasoning~components.}
    \label{fig:fig9}
\end{figure}

The traceability of reasoning through directed acyclic graph (DAG) structures has already been demonstrated in Figure~\ref{fig:fig3}. In~this experiment, we further analyze reasoning visualization based on three fundamental structured inputs from the 30B model test results: target observation priority, GCN transient alerts, and~telescope state information (see Figure~\ref{fig:fig9}). The~results show that, although~more than 7000 distinct reasoning paths can be generated, only 104 atomic reasoning units are activated from the astronomical expert knowledge base. This significantly reduces the complexity of reasoning units while preserving the flexibility of reasoning expression.
Overall, these results indicate that the proposed reasoning visualization framework effectively constrains reasoning complexity without restricting expressive diversity, thereby enabling practical deployment of traceable reasoning. This further confirms the feasibility of the proposed~framework.

\vspace{-10pt}

\unskip

\subsubsection{Performance Comparison of Scheduling~Methods}


	
{
To evaluate the proposed framework under transparent and reproducible conditions, we constructed a controlled benchmark based on the six scheduling scenarios defined in Section~\ref{sect:3.2}. The~benchmark was built from historical real-alert candidates rather than from a purely synthetic target pool. Candidate targets were initialized from ZTF alert records and retained basic alert attributes such as sky position, brightness, and~detection history, consistent with the alert-distribution context of ZTF data products
} 
\citep{2019PASP..131a8002B,2019PASP..131a8001P}. 
{On this basis, the~current observing plan, environmental conditions, and~representative interruption relationships are introduced in a controlled manner to enable reproducible cross-method comparison. In~the reported comparison, each scenario contributes 60 case-level evaluations, yielding 360 evaluations in total for each method. This benchmark should therefore be interpreted as a controlled scheduling benchmark constructed from historical real-alert candidates, rather than as a full replay of a historical observing night or as a complete operational deployment study.}

{The comparison includes representative deterministic schedulers from several common methodological families in astronomical scheduling, together with four Qwen3-based configurations. Specifically, we consider a science-first greedy scheduler} \cite{2024RAA....24a5003C}, { a deadline-oriented rule scheduler, and~a single-step integer-linear-programming (ILP) dispatch scheduler} \cite{2024AJ....167...33H,2025RAA....25a5008J,2026arXiv260108912W}. {These baselines are intended as representative comparators of heuristic, rule-based, and~constrained-optimization approaches that have been widely discussed in the astronomical scheduling literature} \cite{2019AJ....157..151N,2015arXiv150307170L,2026arXiv260414151G}.  {The Qwen3 configurations are \texttt{Qwen3 Only}, \texttt{Qwen3 + Validation}, \texttt{Qwen3 + Validation + Feedback}, and~\texttt{Qwen3 + Framework}. In~Panel A, we report \texttt{Qwen3 Only} and \texttt{Qwen3 + Framework} as the main before/after comparison, while Panel B resolves the intermediate contributions of validation and feedback.}

{Instead of defining a unique ground-truth action for each case, we evaluate all methods against an independently specified reference decision set. This choice reflects the fact that constrained ToO scheduling is generally multi-objective and need not admit a unique canonical action even when the feasible set is well defined. For~each case $i$, let $\mathcal{F}_i$ denote the set of externally executable actions under explicit hard constraints, including instrument status, cloud threshold, target visibility, valid time window, exposure duration, filter validity, and~basic parameter validity. Let $\mathcal{R}_i \subseteq \mathcal{F}_i$ denote the reference decision set obtained by restricting candidates to the highest feasible priority tier and then retaining the Pareto-efficient actions under operational attributes including latest feasible start time, filter-switch cost, motion overhead, and~environment-related risk. The~continuation action (\texttt{{wait}
}/continue the current plan) is treated as admissible only when the current executable plan is not clearly dominated by a feasible interruption option. This construction decouples the evaluation rule from the internal framework logic and is intended to reduce, rather than eliminate, the~circularity associated with a framework-defined success criterion.}

{
We report three complementary metrics. Let $a_i$ be the action produced by a method for case $i$, and~let $N$ be the total number of benchmark cases. The~\emph{{External Executability Rate}} (EER) is defined as }
\begin{equation}
	\mathrm{EER} = \frac{1}{N}\sum_{i=1}^{N}\mathbf{1}\!\left[a_i \in \mathcal{F}_i\right].
\end{equation}

{{The} \emph{{Reference Policy Acceptability}} (RPA) is defined as}
\begin{equation}
	\mathrm{RPA} = \frac{1}{N}\sum_{i=1}^{N}\mathbf{1}\!\left[a_i \in \mathcal{R}_i\right].
\end{equation}

{{Here}, RPA does not denote agreement with a unique ground-truth action. Instead, it measures whether the selected action belongs to the independently specified reference decision set.
To measure ranking quality within the same external decision rule, let \linebreak  $\Pi_i=(b_{i,1},\dots,b_{i,K_i})$ denote the ordered list of rankable reference actions for case $i$. The~case-wise \emph{{Reference Rank Score}} is}

\begin{equation}
	r_i(a_i) =
	\begin{cases}
		0, & a_i \notin \Pi_i,\\
		1, & K_i = 1 \text{ and } a_i = b_{i,1},\\
		1 - \dfrac{k-1}{K_i-1}, & a_i = b_{i,k},\; K_i > 1,
	\end{cases}
\end{equation}
{and the overall \emph{{Reference Rank Score}} (RRS) is}
\begin{equation}
	\mathrm{RRS} = \frac{1}{N}\sum_{i=1}^{N} r_i(a_i).
\end{equation}

{For the overall metrics, we report 95\% confidence intervals using Wilson intervals for binary rates and bootstrap intervals for mean ranking scores.} 
In Table~\ref{tab:table3},{scenario-wise comparisons are reported using RPA, while RRS is retained as the overall ranking-quality metric. This presentation keeps the scenario-level interpretation aligned with externally admissible decision selection under the independently specified reference set.}

 Table~\ref{tab:table3} {reports both the main method comparison and the internal ablation of the proposed pipeline over 360 case-level evaluations per method. For~readability, the~main table reports representative baselines; additional deterministic baselines that saturate at $1.000$ on several aggregated metrics are omitted from Panel A because they add little discrimination under the simpler control scenarios. Pure Qwen3 performs substantially worse than the deterministic baselines, with~an overall $\mathrm{EER}=0.692$, $\mathrm{RPA} =0.531$, and~$\mathrm{RRS}=0.612$. By~contrast, \texttt{Qwen3 + Framework} reaches $\mathrm{EER}=1.000$, $\mathrm{RPA}=0.944$, and~$\mathrm{RRS}=0.983$.}

 {Scenario-wise results further clarify the behavior of the framework. The~most difficult cases are $S3$ and $S5$, where several feasible interruption choices or urgent follow-up opportunities must be distinguished under coupled operational trade-offs. In~these scenarios, \texttt{Qwen3 Only} attains RPA values of only $0.050$ and $0.050$, whereas \texttt{Qwen3 + Framework} increases them to $0.733$ and $0.933$, respectively. By~contrast, performance differences are much smaller in the relatively clean control scenarios $S1$, $S2$, and~$S6$, where most methods already satisfy the external decision criteria.} {The comparison also helps calibrate a reasonable expectation for AI scheduling relative to deterministic methods. As~a stand-alone scheduler, pure Qwen3 is not competitive with strong constraint-aware baselines under explicit constraints. For~example, in~the difficult S5 scenario, pure Qwen3 attains an RPA of $0.050$, compared with $0.483$ for the greedy baseline and $1.000$ for the ILP baseline, whereas the full framework increases the AI-assisted configuration to $0.933$. By~contrast, S6 is more informative as a safety stress case than as a discriminative policy-ranking case, because~once physically invalid inputs are recognized, the~admissible response often collapses to safe rejection, repair, or~deferral.}

{The ablation results show that the improvement arises in stages. Validation alone fully restores executability ($\mathrm{EER}=1.000$), but~is not sufficient to recover decision quality in difficult cases ($\mathrm{RPA}=0.683$, $\mathrm{RRS}=0.765$). The~largest gain appears after feedback-based regeneration, with~\texttt{Qwen3 + Validation + Feedback} reaching $\mathrm{RPA}=0.928$ and \mbox{$\mathrm{RRS}=0.978$}. Adding the full reasoning-audit layer yields a further improvement to \mbox{$\mathrm{RPA}=0.944$} and $\mathrm{RRS}=0.983$, while also introducing explicit audit flags and traceability~hooks.}

{Overall, the~results indicate that the proposed framework substantially improves executability and reliability relative to pure AI generation, especially in the more difficult multi-constraint interruption scenarios. Under~this controlled benchmark, the~framework-enhanced AI configuration approaches the stronger deterministic baselines in aggregate decision quality, although~it does not surpass the strongest deterministic baselines in clean benchmark conditions. We emphasize that the primary goal of our framework is to ensure the executability and correctness of AI‑generated decisions; achieving optimal scheduling performance (e.g., surpassing deterministic baselines in clean scenarios) remains a matter of model capability and task‑specific tuning, which is beyond the scope of this work.}


\subsubsection{Comprehensive Analysis of the Validation~Framework}

After introducing the hierarchical validation framework, the~executability of AI-generated scheduling proposals remains stable across all test scenarios. The~multi-level validation mechanism effectively identifies and resolves constraint conflicts and logical inconsistencies in model outputs, thereby reducing the occurrence of non-executable decisions and improving the reliability of scheduling results.At the reasoning level, the~introduction of atomic reasoning units and structured reasoning representation enables traceability and reuse of the reasoning process. This approach maintains expressive flexibility while effectively controlling reasoning complexity, providing support for interpretable scheduling decisions.In scheduling scenarios constructed from historical ZTF transient alerts, the~experimental results further show that, under~complex constraints, pure AI-based scheduling struggles to consistently satisfy multi-target and multi-constraint requirements. In~contrast, the~proposed multi-level validation framework improves decision consistency and overall scheduling quality, bringing performance close to that of advanced traditional scheduling methods.
Overall, the~results demonstrate the effectiveness of the proposed framework from three aspects: reliability, interpretability, and~practical scheduling performance. By~introducing structured validation and feedback mechanisms, the~framework mitigates the uncertainty in AI reasoning and enables AI-based scheduling methods to achieve stability comparable to traditional approaches while preserving their flexibility, thereby supporting their application in real telescope observation~scheduling.

\section{Discussion}

The experimental results reveal two key problems in the application of large AI models to astronomical scheduling. First, ``hallucinations'' in AI outputs reduce reliability, and~the AI-generated scheduling decisions still require manual verification. 
{This efficiency bottleneck remains a challenge for deploying LLM-based approaches in high-reliability settings, motivating the validation framework proposed in this work.}
Second, the~reasoning logic and process cannot be explicitly represented, and~the reasoning expressions are not unique. Given the same input, the~output is difficult to predict. {The experimental results show that the proposed framework can effectively reduce the uncertainty of AI-generated scheduling decisions under complex constraints, thereby improving the executability and reliability of AI-assisted scheduling.}

Further analysis indicates that the effectiveness of the framework mainly arises from early-stage filtering of erroneous decisions and validation of constraint consistency. Through multi-level verification of data references, logical consistency, and~observational and instrumental constraints, non-executable or inconsistent scheduling proposals can be identified and blocked before entering the execution stage, thereby reducing the impact of erroneous decisions on observational resources.  {A reasonable expectation for current LLM-based schedulers is therefore not that they outperform strong deterministic schedulers in clean and fully specified settings, where classical methods remain more efficient. Rather, their practical role is to provide flexible proposal generation or prioritization in noisy, heterogeneous ToO environments, provided that an independent validation layer enforces executability and policy safety.}

{Unlike structured-agent pipelines that rely on the LLM itself for iterative self-verification, our framework performs validation through a decoupled, deterministic layer using explicit database queries and constraint checks. This design ensures a clear boundary between proposal generation and execution safety. Testing the framework with more capable frontier models is planned for future work.}
{This assessment is grounded in the framework's core strength: by making the AI decision process traceable and verifiable through structured reasoning representation and step-level consistency checking, one can credibly evaluate where AI-assisted scheduling is competitive and where it is not.}

{These challenges—opaque reasoning processes, non-traceable decisions, and~the risk of generating non-executable outputs—motivate the two complementary mechanisms introduced in this work.}	
On the one hand, a~hierarchical visual validation framework is proposed. By~combining knowledge retrieval, computation, and~rule-based validation, the~framework performs multi-dimensional verification of model references, astronomical knowledge, and~instrument and environmental states, and~incorporates a feedback mechanism to enable rapid correction of model outputs.
On the other hand, at~the reasoning level, atomic reasoning units and their dependency relationships were introduced. Scheduling decisions are represented as a structured process composed of a series of interconnected reasoning steps. This enables step-by-step verification of the reasoning process and supports error localization as well as post hoc analysis. {Over extended operation, the~accumulating atomic reasoning library and progressively refined validation rules are expected to make the reasoning space increasingly manageable for human audit and systematic reuse.}

The experimental results further show that, while maintaining expressive capability, the~number of reasoning units exhibits a convergence trend and achieves a high level of reusability. This helps control reasoning complexity and makes the reasoning process practically manageable. These results provide a technical foundation for introducing AI methods into high-reliability astronomical scheduling scenarios. Experiments conducted on models with up to 30B parameters applied to telescope scheduling demonstrate that the proposed validation framework enables visualization and reuse of model outputs. Across more than 2000 test cases, the~average repetition rate of reasoning logic exceeds 80\%, while the number of reusable atomic reasoning units remains below 500, indicating that the complexity has entered a range acceptable to human users, thereby supporting high-reliability scheduling requirements.
At the same time, the~framework shows good extensibility and provides a new approach for reliability validation in AI applications beyond astronomy. It can be further extended to other high-reliability domains, such as power system energy management and medical AI~diagnosis.

There are still aspects that require further investigation. For~example, the~balance between the granularity of AI reasoning logic and the efficiency of practical validation needs further optimization. 
{Several limitations of this study point to directions for future work. Future work will validate the stability and adaptability of this framework in real observational systems under long-term operational conditions. In~addition, this work focuses on dynamic ToO scheduling scenarios and does not evaluate static offline nightly scheduling, where classical constrained optimization methods are well established; end‑to‑end simulation evaluation is left for future work. The~inference times reported in this study were obtained using a an NVIDIA GeForce RTX 4070 GPU (NVIDIA Corporation, Santa Clara, CA, USA) and reflect current experimental conditions; a comparative cost analysis between traditional and LLM-based scheduling under realistic observatory workloads is planned for future work. Our contribution is improving AI reliability under incomplete/noisy inputs.}

\vspace{6pt}
\authorcontributions{Conceptualization, C.W. and H.X.; methodology, H.X.; software, H.X.; validation, H.X. and C.W.; formal analysis, H.X.; investigation, H.X.; resources, C.W.; data curation, H.X.; writing—original draft preparation, H.X.; writing—review and editing, C.W.; visualization, H.X.; supervision, C.W.; project administration, C.W.; funding acquisition, C.W. All authors have read and agreed to the published version of the~manuscript.}

\funding{{This} 
 research was funded by the National Key R\&D Program of China, grant number 2023YFA1608300; and the Yunnan Revitalization Talent Support Program (Young Talent Project).}

\dataavailability{\textls[-15]{The code developed for this study is publicly available on {GitHub} 
 {at} 
 \url{https://github.com/xiaohengchu24/A-Hierarchical-Validation-and-Traceable-Reasoning-Framework-Code}}  ({accessed on)}
. A~frozen version has been archived in Zenodo with DOI: 10.5281/zenodo.19738938. The~ZTF transient alert data used in this study are publicly available.}

\acknowledgments{This study makes use of two large language models: the Qwen3 model developed by the Qwen Team~\cite{qwen3technicalreport} and the Phi-4 model developed by Microsoft Research~\cite{abdin2024phi4technicalreport}.
}

\conflictsofinterest{The authors declare no conflicts of~interest.}

\begin{adjustwidth}{-\extralength}{0cm}

\reftitle{References}

\PublishersNote{}

\end{adjustwidth}
\end{document}